\documentclass[conference]{IEEEtran}

\usepackage{cite}
\usepackage{capt-of}
\usepackage{amsmath}
\usepackage{amssymb}

\usepackage{graphicx}
\usepackage{booktabs}
\usepackage{multirow}

\usepackage{xcolor}
\usepackage{url}

\usepackage{balance}

\usepackage{hyperref} 
\usepackage{graphicx}
\usepackage{amsmath}
\usepackage{amsthm}
\usepackage{subfigure}
\usepackage{balance}
\usepackage{multirow}
\usepackage{xspace}
\usepackage{enumitem}

\newcommand{\squishlist}{
 \begin{list}{$\bullet$}
  {  \setlength{\itemsep}{0pt}
     \setlength{\parsep}{3pt}
     \setlength{\topsep}{3pt}
     \setlength{\partopsep}{0pt}
     \setlength{\leftmargin}{2em}
     \setlength{\labelwidth}{1.5em}
     \setlength{\labelsep}{0.5em}
} }
\newcommand{\squishlisttight}{
 \begin{list}{$\bullet$}
  { \setlength{\itemsep}{0pt}
    \setlength{\parsep}{0pt}
    \setlength{\topsep}{0pt}
    \setlength{\partopsep}{0pt}
    \setlength{\leftmargin}{2em}
    \setlength{\labelwidth}{1.5em}
    \setlength{\labelsep}{0.5em}
} }

\newcommand{\squishdesc}{
 \begin{list}{}
  {  \setlength{\itemsep}{0pt}
     \setlength{\parsep}{3pt}
     \setlength{\topsep}{3pt}
     \setlength{\partopsep}{0pt}
     \setlength{\leftmargin}{1em}
     \setlength{\labelwidth}{1.5em}
     \setlength{\labelsep}{0.5em}
} }

\newcommand{\squishend}{
  \end{list}
}

\newcommand{\eat}[1]{}

\newcommand{\kw}[1]{{\ensuremath{\mathsf{#1}}}}

\newcommand{\sstab}{\rule{0pt}{8pt}\\[-2.4ex]}
\newcounter{ccc}

\newcommand\redout{\bgroup\markoverwith
{\textcolor{red}{\rule[.5ex]{2pt}{2pt}}}\ULon}

\newcommand{\M}{{\mathcal M}}
\newcommand{\C}{{\mathcal C}}
\newcommand{\gnn}{\kw{GNN}}
\newcommand{\gnns}{\kw{GNNs}}

\newcommand{\sparql}{\kw{SPARQL}}
\newcommand{\slice}{\kw{SliceGX}}
\newcommand{\sliceq}{\kw{SliceGXQ}}

\begin{document}

\title{Declarative Explanations for Graph Neural Networks:
A Demonstration}


\author{
\IEEEauthorblockN{
Cibo Yu\IEEEauthorrefmark{1},
Haitong Tang\IEEEauthorrefmark{1},
Yinghui Wu\IEEEauthorrefmark{2},
Arijit Khan\IEEEauthorrefmark{3},
Tingting Zhu\IEEEauthorrefmark{1},
Tingyang Chen\IEEEauthorrefmark{1},
and Xiangyu Ke\IEEEauthorrefmark{1}
}

\IEEEauthorblockA{
\IEEEauthorrefmark{1}
Zhejiang University, Hangzhou, China\\
\{cbyu, tht, tingtingzhu, chenty, xiangyu.ke\}@zju.edu.cn
}

\IEEEauthorblockA{
\IEEEauthorrefmark{2}
Case Western Reserve University, Cleveland, USA\\
yxw1650@case.edu
}

\IEEEauthorblockA{
\IEEEauthorrefmark{3}
Bowling Green State University, Bowling Green, USA\\
arijitk@bgsu.edu
}
}

\maketitle


\begin{abstract}
Graph Neural Networks (\gnns) have proven effective for graph
analytical tasks, yet their ``black-box'' nature complicates the
understanding of their layer-wise behavior and hinders their
application to large graphs.
We demonstrate \sliceq, an end-to-end declarative and interactive
system for querying and exploring layer-wise explanations for
\gnns\ over large-scale graphs.
In particular, \sliceq\ provides
\textbf{(1)} an interactive graphical user interface for dataset,
model, and explanation configuration;
\textbf{(2)} a lightweight \textsf{SPARQL}-like parser that allows
users to specify target nodes, layers, explanation modes, and
constraints; and
\textbf{(3)} an integrated execution and visualization workflow
that invokes \textsf{SliceGX} \cite{zhu2026slicegxlayerwisegnnexplanation} as its underlying explanation engine.
We demonstrate how users can diagnose a misclassified node,
identify problematic layers and influential messages, and perform
targeted model debugging, together with large-scale graph
interpretation and fraud-detection analysis.
Our demonstration video and source code are available at
\url{https://github.com/jinzhaoqet/SliceGXQ}.
\end{abstract}




\section{Introduction}
\label{sec:intro}
Graph Neural Networks (\gnns) have demonstrated remarkable success across various tasks, including node classification. 
By leveraging message passing and node feature aggregation, \gnns~ effectively learn node representations by integrating graph structural information with node features. 
This capability has enabled the widespread applications of \gnns\ in domains such as social network analysis, 
cyber threat analysis, bioinformatics, and recommendation systems.

Due to complex message-passing mechanisms and end-to-end learning, \gnns\ are often queried and accessed 
as ``black-boxes'', making their decision-making process opaque to users. Recent efforts in explainable graph learning~\cite{10.1145/3749116.3749122} have developed various GNN explainers e.g., \cite{ying2019gnnexplainer, yuan2021explainability, YuanTHJ20}. These methods identify essential substructures and data features responsible for \gnns' outputs, and their influence and impact. 

Nevertheless, existing \gnn\ explanation techniques exhibit several limitations \cite{0001K0B26, 10.1145/3626246.3654735}:
{\bf (1)} Most of them provide {\em static, one-time} explanations that focus solely on the final outputs rather than insights into how a \gnn\ processes information progressively. This lack of fine-grained mechanisms for interpretation and manipulation hinders users' ability to localize and explore the internal reasoning behavior of \gnns. 
{\bf (2)} Many explainers are themselves ``black-box'', making it difficult to validate their outputs; 
and {\bf (3)} Users often need to tailor explanations to specific requirements, such as filtering for particular substructures in the output or focusing on specific layers of the model, which typically requires manual intervention and domain expertise. 
Hence, a \textit{declarative} explanation system is essential for users to efficiently specify 
queries and easily customize explanations according to their needs.

To this end, we introduce a declarative framework that supports a class of \sparql-like explanatory queries, called \slice\ queries,  
to support layer-wise explanations\footnote{We present layer-wise \gnn\ explanations in the full research paper~\cite{zhu2026slicegxlayerwisegnnexplanation}.} of \gnns. 
A \sliceq\ query $Q$ follows a general structured format: 

\begin{center}
\fbox{
\begin{minipage}{.25\textwidth}
\kw{\textbf{EXPLAIN}} ($\M$, $V_T$) \kw{\textbf{AT}} $l_t$ \\
\kw{\textbf{FROM}}$\{G, \L\}$ \\
\kw{\textbf{WHERE}} {\em P, k} \kw{\textbf{WITH}} {\em mode} 
\end{minipage}
}
\end{center}

Specifically, the following declarative operators are integral to \sliceq\ queries: 
\textbf{(1) EXPLAIN} specifies the primary objective of the query, which is to generate explanations for a pre-trained \gnn\ model ($\M$) and test nodes of interest ($V_T$); 
{\textbf{(2) AT} allows users to inspect a specific layer ($l_t$) within the \gnn\ to interact with,
enabling fine-grained, progressive analysis; 
\textbf{(3) FROM} defines the graph dataset ($G$) and selected layers ($\L$) to generate explanations for model behaviors at $l_t$;
\textbf{(4) WHERE} applies ``shape'' constraints on the explanations as subgraphs, such as filtering by a {\em graph/path pattern} ($P$), as seen in 
SPARQL queries; or limiting the result size ($k$); 
and \textbf{(5) WITH} specifies the explanation mode (e.g., model diagnosis vs. interpretation, as detailed in \S\ref{sec:core}). Moreover, \sliceq\ supports SQL-like aggregators, such as ORDER BY, allowing users to sort the generated explanations by specific utility.WHERE supports graph/path-pattern constraints, \(k\) bounds the explanation size, and ORDER BY ranks explanations by their explainability scores.

\sliceq\ addresses critical challenges in \gnn\ explanation, including limited granularity, interactivity, and customization by introducing a declarative paradigm that allows users to access layer-wise explanations through \sparql-like queries. 
Different from \textsf{SliceGX} \cite{zhu2026slicegxlayerwisegnnexplanation}, \sliceq\ focuses on the declarative query interface, interactive visualization, and end-to-end explanation workflow built upon the \textsf{SliceGX} engine.The key contributions of \sliceq\ are summarized as follows.
\begin{itemize}[leftmargin=*]
\item \sliceq\ enables a declarative workflow to generate 
explanations throughout the decision-making process of \gnns{} at any specific, segmented layers, and allows users to 
construct flexible, configurable queries tailored to application needs.
\item \sliceq\ contains an 
interactive
visualization toolkit for the configuration 
of explanation context, inference-time explanation browsing, and statistical analysis. This interactive interface empowers users to gain finer-grained layer-wise insights into both data properties and model behavior.
\item We demonstrate the effectiveness of \sliceq\ through
layer-wise model diagnosis and debugging scenarios. 
By visualizing how the model processes information at each layer, users can identify potential error sources and refine the model accordingly. 
\end{itemize}
%


\section{System Overview}
\label{sec:system}
The workflow of \sliceq\ is shown in Figure~\ref{fig:architecture}.
The front end consists of four modules: Graph Data Selection,
\gnn\ Model Selection, Query-Based Explanations, and Interactive
Visualization.
The Query-Based Explanations module connects the front end with
four back-end modules: Query Interpretation, Model Slicing,
Explanation Subgraph Finding, and Verification Operator.
These modules process user configurations, generate layer-wise
explanations, and return verified results for interactive inspection.

\begin{figure}[tb!]
\centering
\vspace{-3.5mm}
\centerline{\includegraphics[width=\columnwidth]{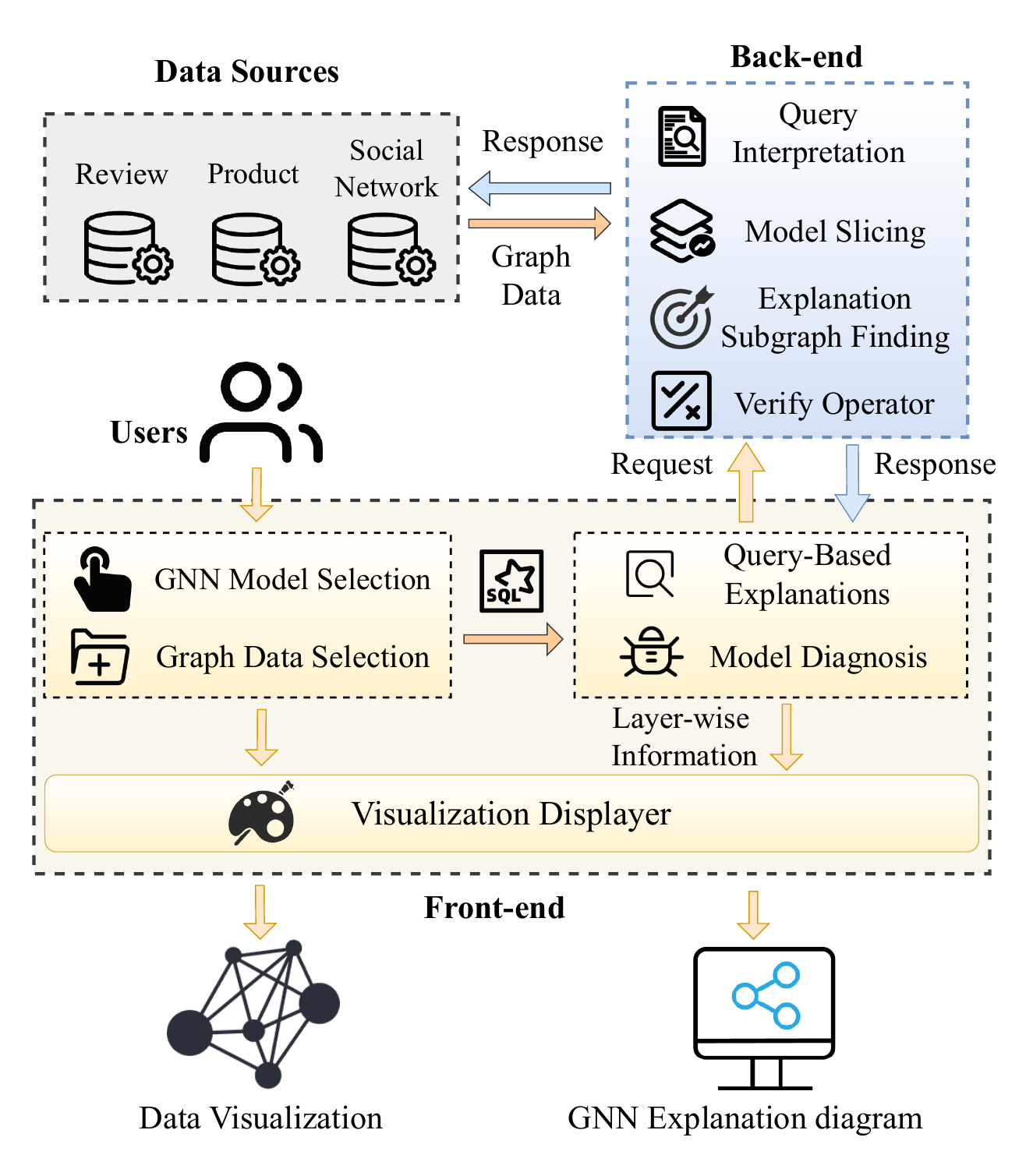}}
\vspace{-5mm}
\caption{\sliceq~workflow. Users select a graph and a
\gnn\ model, submit an explanation request, and inspect the
layer-wise results returned. }
\vspace{-5.5mm}
\label{fig:architecture}
\end{figure}

\vspace{0.5mm}
\noindent\textbf{Configuration:}
The Graph Data Selection module allows users to select a dataset
and specify target nodes.
The \gnn\ Model Selection module supports model and architecture
configuration.
These selections are passed to the Query-Based Explanations module,
where users specify the target layer, explanation mode, size bound,
and optional structural constraints.
The results are displayed through Interactive Visualization.

\vspace{0.5mm}
\noindent\textbf{Query-Based Explanations:}
This module supports Configuration Mode for novice users and
Query Mode for advanced users.
The former provides a graphical form, whereas the latter supports
customized SPARQL-like statements.
Both modes share the same query representation and execution
pipeline.
Representative statements are shown in
Figure~\ref{fig:input_examples}.
The back-end modules are executed as follows.

\sstab
\textbf{(1) Query Interpretation}
parses and validates a query and translates it into an executable
task.
Invalid datasets, models, nodes, layers, or constraints produce an
error message before explanation generation.
A valid query is translated into a configuration 
$\C=(G,\M,V_T,\L,l_t,P,k,\mathit{mode})$,
where $G$ specifies the graph, $\M$ is a selected \gnn, $V_T$ is the
target-node set, $\L$ contains the source layers, $l_t$ is the
target layer, $P$ specifies structural constraints, $k$ bounds the
explanation size, and $\mathit{mode}$ allows users to switch between ``interpretation'' or
``diagnosis'' modes, which 
localizes explanations responsible 
for correct or incorrect outputs, respectively.

\sstab
\textbf{(2) Model Slicing}
constructs a layer-specific sliced model from the selected \gnn\
and target layer.
It preserves the computation up to that layer and provides the
intermediate representations required for explanation generation.

\sstab
\textbf{(3) Explanation Subgraph Finding}
invokes the underlying \textsf{SliceGX} procedure and employs a
greedy strategy to identify high-quality explanatory nodes.
It constructs a connected explanation subgraph that satisfies the
specified size and structural constraints.

\sstab
\textbf{(4) Verifier}
checks whether a candidate explanation satisfies the factual or
counterfactual properties defined in \S\ref{sec:core} for 
the underlying configuration. 
Only verified explanations are returned for layer-wise comparison and visual analysis. 

\vspace{0.5mm}
\noindent\textbf{Model Diagnosis and Debugging:}
Users can inspect the explanations to identify layers or nodes
associated with incorrect predictions.
They may then initiate layer fine-tuning or architecture
optimization and compare the model behavior before and after
debugging.
These operations are user-initiated rather than automatically
applied by the explanation engine.
\begin{figure*}[!t]
    \centering

    \begin{minipage}[t]{0.48\textwidth}
        \centering
        \includegraphics[
            height=4.2cm,
            keepaspectratio
        ]{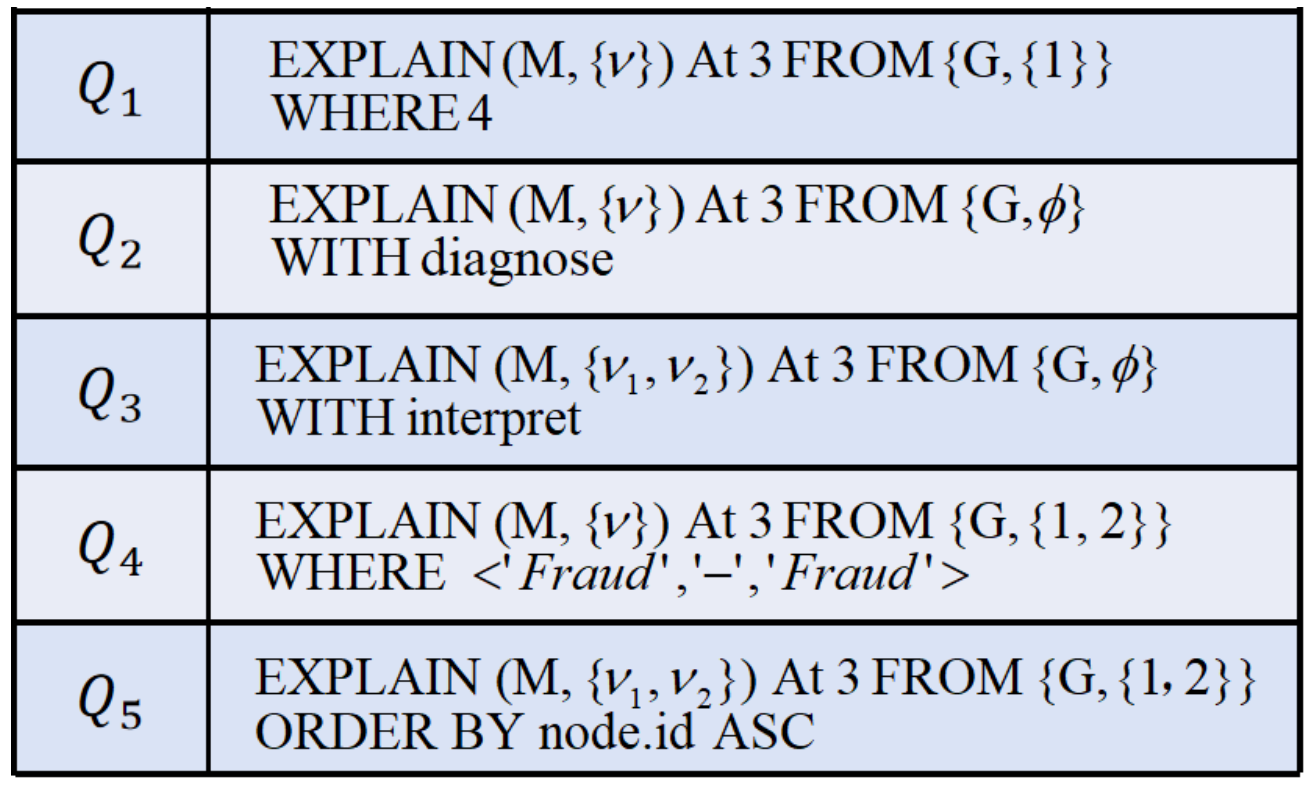}
        \vspace{-2mm}
        \captionof{figure}{Examples of SPARQL-like \sliceq\
        statements.}
        \label{fig:input_examples}
    \end{minipage}
    \hfill
    \begin{minipage}[t]{0.48\textwidth}
        \centering
        \includegraphics[
            height=4.2cm,
            keepaspectratio
        ]{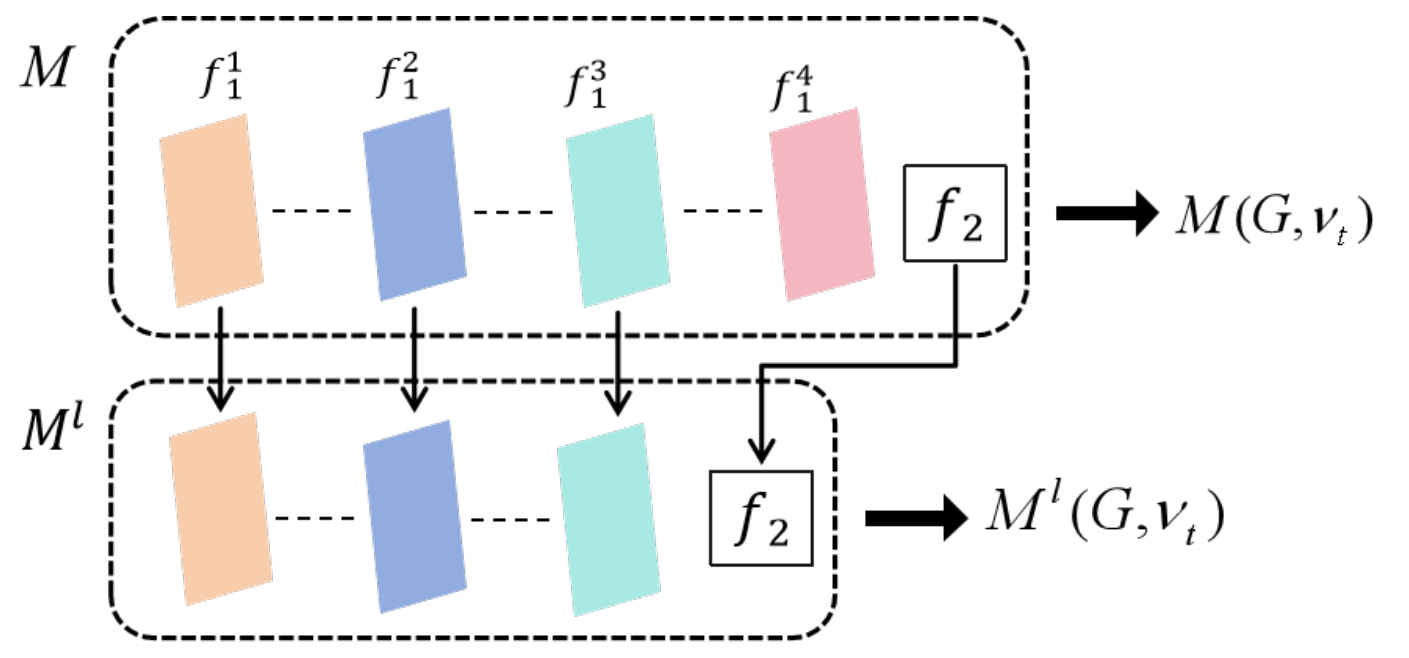}
        \vspace{-6mm}
        \captionof{figure}{An example of a sliced model for $l=3$.}
        \label{fig:slicedmodel}
    \end{minipage}

    \vspace{-6mm}
\end{figure*}


\section{Explanation Engine}
\label{sec:core}

We briefly introduce the \textsf{SliceGX} explanation engine
underlying \sliceq; further algorithmic details are provided
in~\cite{zhu2026slicegxlayerwisegnnexplanation}.
Given a pre-trained \gnn\ model $\M$, a graph $G$, and a target
node $v_t$ whose output $\M(G,v_t)$ is to be explained, the engine
returns a set of \emph{explanatory nodes} that clarify the model
behavior at a selected layer, together with a connector set $V_c$
that preserves the connectivity of the explanation subgraph.

\vspace{0.5mm}
\noindent\textbf{Model Slicing.}
Given a \gnn\ model $\M$ with an $L$-layer encoder $f_1$ and a
predictor $f_2$, the $l$-sliced model $\M^l$, for
$l\in[1,L]$, retains the first $l$ encoder layers and feeds their
output embeddings to the original predictor.
As illustrated in Figure~\ref{fig:slicedmodel}, this construction
allows users to inspect the model behavior at a particular layer
without retraining the model.
Because the sliced and original models may exhibit inconsistent
behavior, the verification operator checks the reliability of the
generated explanations.

\vspace{0.5mm}
\noindent\textbf{Verify Operator.}
For a target layer $l$, the operator checks an explanation subgraph
$G_s^l\subseteq G$ against both the original model $\M$ and the
sliced model $\M^l$.
A factual explanation preserves the relevant predictions when the
models are evaluated on $G_s^l$, whereas a counterfactual
explanation identifies a subgraph whose removal changes the
relevant prediction.
The operator also checks that the original and sliced models are
behaviorally consistent on the full graph.
Only explanation subgraphs satisfying the requested factual or
counterfactual condition are returned by \sliceq. A quantitative analysis of verification rejection rates is left for future evaluation.

\vspace{0.5mm}
\noindent\textbf{Explanation Generation.}
The underlying engine employs a greedy strategy to construct a
high-quality explanation subgraph for $\M(G,v_t)$ at each selected
layer.
It uses a bi-criteria explainability score that combines
\emph{relative influence} and \emph{embedding diversity}.

Given an explanation subgraph $G_s^l$ with explanatory-node set
$V_s^l$, its explainability at layer $l$ is defined as
\begin{equation}
\label{quality}
f(G_s^l)=\gamma I(V_s^l)+(1-\gamma)D(V_s^l),
\end{equation}
where $I(V_s^l)$ measures the relative influence of the explanatory
nodes, $D(V_s^l)$ measures their embedding diversity, and
$\gamma\in[0,1]$ balances the two criteria.
This score captures layer-specific intermediate representations.

\vspace{0.5mm}
\noindent\textbf{Diagnosis and Interpretation Modes.}
\sliceq\ exposes two explanation modes through its declarative
interface.
\emph{Diagnose Mode} identifies layers and explanatory nodes
associated with an incorrect prediction, whereas
\emph{Interpret Mode} inspects how the model's internal
representations evolve across layers.

\vspace{-1mm}
\section{Demonstration Scenarios}
\label{sec:demonstration}

We demonstrate \sliceq\ through an end-to-end workflow covering
dataset/model selection, declarative queries, layer-wise explanation,
model diagnosis, and user-initiated debugging.
For each scenario, we describe the analysis question, system
operations, and returned results.

\begin{figure*}[t]
\vspace{-1mm}
\centering
\centerline{\includegraphics[width=\textwidth]
{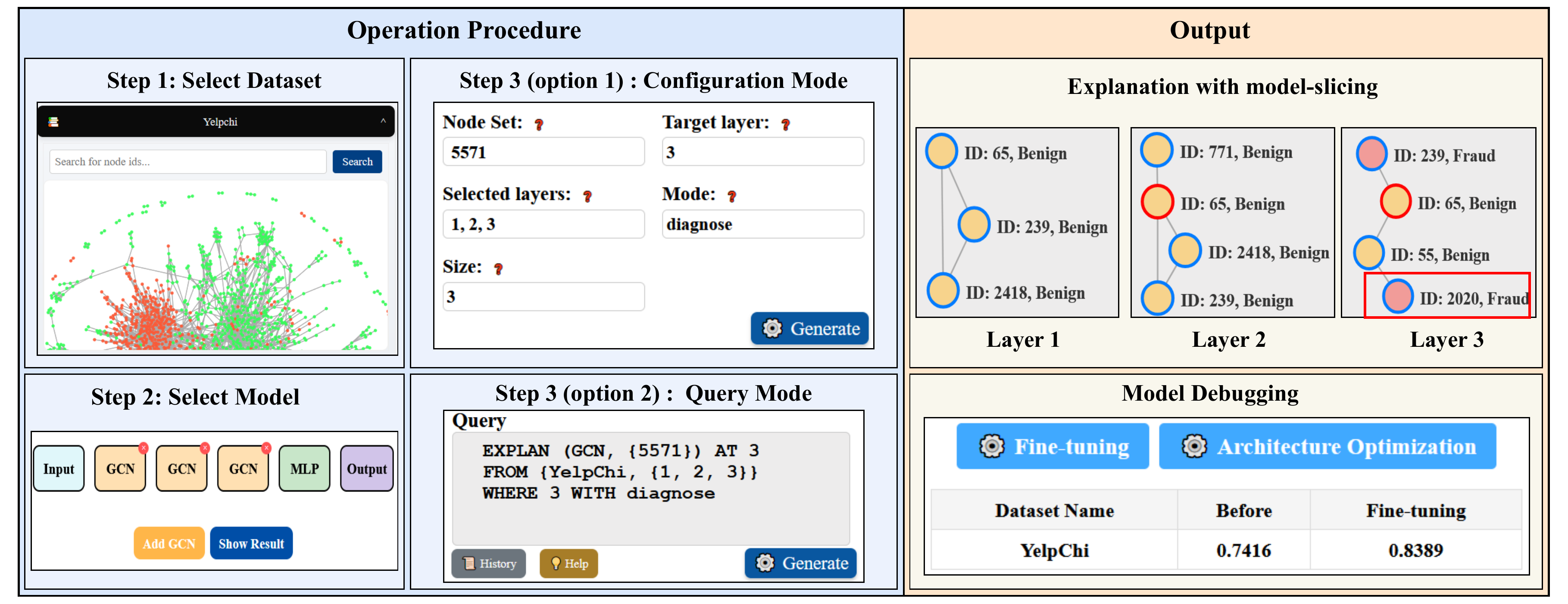}}
\vspace{-3.5mm}
\caption{\small End-to-end workflow of \sliceq.
The left panels show dataset and model selection and the two query
input modes; the right panels show layer-wise explanation results
and user-initiated model debugging.
In the output panel, red and yellow nodes indicate predictions of
``Fraud'' and ``Benign'', respectively; blue-bordered and
red-bordered nodes are explanatory and connector nodes, respectively.}
\vspace{-4mm}
\label{fig:screenshot}
\end{figure*}

\vspace{0.5mm}
\noindent\textbf{Scenario 1: Workflow Walkthrough.}
As illustrated in Figure~\ref{fig:screenshot}, a user completes an
explanation task in three steps.
\textbf{Step 1: Dataset Selection.}
The user selects a graph dataset, inspects its node and edge
statistics, and specifies target nodes by entering their identifiers
or clicking them in the graph view.
\textbf{Step 2: Model Selection.}
The user selects a pre-trained \gnn\ model, inspects its architecture,
and browses its prediction results.
\textbf{Step 3: SliceGXQ Query.}
A novice user specifies the target layer, explanation mode, size
bound, and constraints through Configuration Mode, whereas an
advanced user directly submits a SPARQL-like query through Query
Mode.
Both modes share the same back-end pipeline, which returns
visualized layer-wise subgraphs for inspection and follow-up queries.

\vspace{0.5mm}
\noindent\textbf{Scenario 2: Model Diagnosis.}
We consider a fraud-detection task on the \textsf{YelpChi} graph,
where the \gnn\ distinguishes ``Benign'' and ``Fraud'' instances.
The user's goal is to determine how the returned explanation
changes across layers and which layer-specific nodes should be
inspected for diagnosing the target prediction.

In the example shown in Figure~\ref{fig:screenshot}, the user
selects node 5571 as the query target, layers 1--3 as the source
layers, layer 3 as the target layer, a size bound of 3, and Diagnose
Mode.
This request follows the diagnosis-query pattern represented by
$Q_2$ in Figure~\ref{fig:input_examples}.
After the query is submitted, \sliceq\ performs query
interpretation, model slicing, explanation-subgraph generation, and
verification, and then displays the layer-wise explanation
subgraphs.

The displayed class of node 239 remains ``Benign'' in the
explanations for layers 1 and 2 but changes to ``Fraud'' when the
third layer is included.
The layer-3 result also contains node 2020, a ``Fraud''-labeled
connector node that links the returned explanation structure.
Because the returned explanation exhibits this change only when
the third layer is included, the user treats the third-layer
aggregation as the primary diagnosis target.
The user then inspects nodes 239 and 2020 and their relationships,
narrowing the analysis to the third layer and a compact node set.
Rather than directly attributing the prediction error to a single
node, the diagnosis prioritizes a small set of layer-specific
structures for subsequent user inspection and debugging.

\vspace{0.5mm}
\noindent\textbf{Scenario 3: Scalability of Explanation Generation.}
We next evaluate progressive explanation generation on two graph
scales.
\textbf{Coauthor CS} is a co-authorship benchmark containing
18\,333 nodes and 163\,788 edges, while \textbf{Amazon} is a
million-node product graph containing 1\,569\,960 nodes and
264\,339\,468 edges.
The analysis question is whether \sliceq\ can generate useful
layer-wise explanations as the graph and the model's receptive
field become larger.

We issue interpretation queries for layers 1--3, following the
query pattern represented by $Q_3$ in
Figure~\ref{fig:input_examples}, and report explanation-generation
time and fidelity in Figure~\ref{fig:fix}.
As the selected layer increases, fidelity improves as more of the
inference process is covered, while generation time increases
moderately; the reported time excludes front-end interaction.

Further algorithmic evaluations are available
in~\cite{zhu2026slicegxlayerwisegnnexplanation}.

\begin{figure}[t]
\centering
\centerline{\includegraphics[width=\columnwidth]{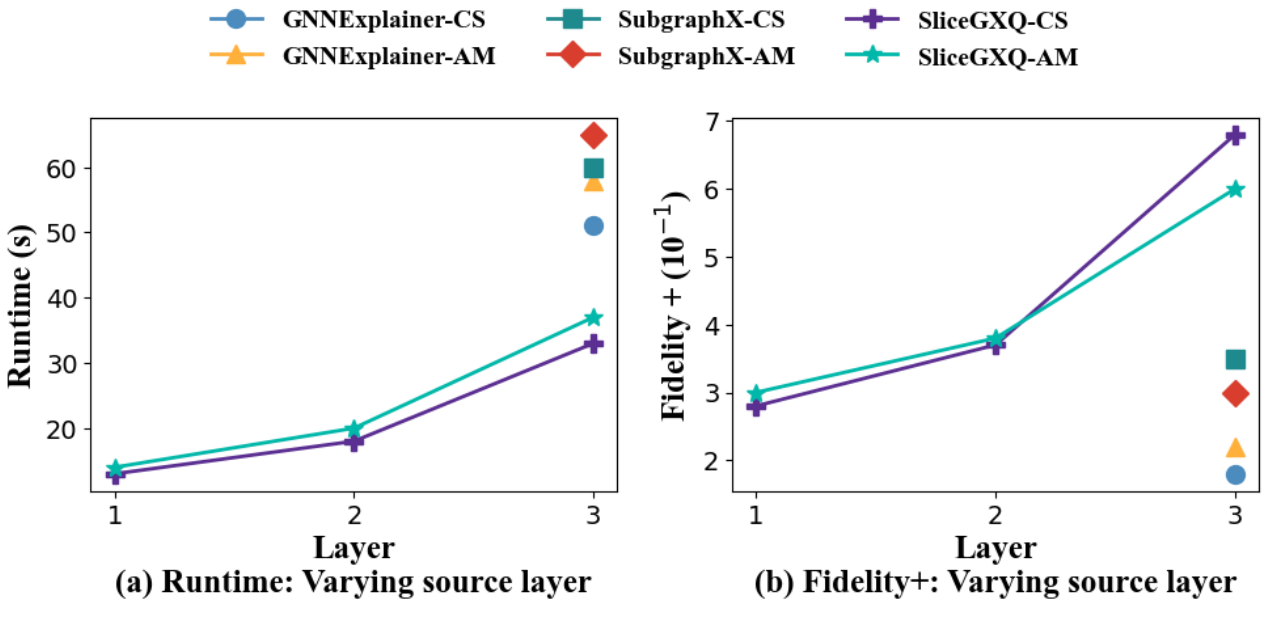}}
\vspace{-4mm}
\caption{\small Explanation-generation time and fidelity of \sliceq\
across GNN layers on an NVIDIA GeForce RTX 3080 GPU.
GNNExplainer and SubgraphX appear only at Layer 3 as final-output explainers.}
\vspace{-6mm}
\label{fig:fix}
\end{figure}

\noindent\textbf{Scenario 4: Model Debugging.}
Scenario 2 identifies the third layer as the primary diagnosis
target based on changes in the returned explanation. We next show
how \sliceq{} helps users assess whether modifying this layer can
improve the model while keeping the remaining layers unchanged.
Based on the diagnosis, the user selects the third layer and
initiates the fine-tuning operation from the debugging panel in
Figure~\ref{fig:screenshot}.

The interface presents the model performance before and after the
operation.
In the shown example, fine-tuning the diagnosed third layer
increases test accuracy from $0.7416$ to $0.8389$, illustrating the workflow rather than a
causal advantage of this layer.
The user can compare the updated performance with the original
result and rerun the same explanation query on the updated model.
The resulting layer-wise explanations allow the user to examine
whether the diagnosed behavior has changed after fine-tuning,
completing an iterative diagnosis-and-debugging loop.
This diagnosis-to-debugging workflow illustrates how declarative
layer-wise explanations can guide a targeted model refinement
decision.

\section{CONCLUSION}

We develop \sliceq, a user-friendly, interactive, and declarative
platform for accessing configurable and queryable \gnn\ explanations
via model slicing and a class of structured explanatory query
operators. Built on \textsf{SliceGX} as its underlying explanation
engine, our demonstration provides human experts with a comprehensive
visual and operational tool for exploring layer-wise explanations
and understanding the model's internal behavior for model
interpretation and diagnosis. Additionally, it allows users to
identify problematic layers and influential explanatory nodes for
targeted model debugging, while configuring and customizing
explanations across diverse application scenarios.

\vspace{-1mm}
\section{AI-Generated Content Acknowledgement}
This paper has no AI-generated content.



\vspace{-1mm}
\bibliographystyle{IEEEtran}
\bibliography{ref}

@inproceedings{zhu2026slicegxlayerwisegnnexplanation,
      title={SliceGX: Layer-wise GNN Explanation with Model-slicing}, 
      author={C. Yu and T. Zhu and T. Chen and Y. Wu and A. Khan and X. Ke},
      year={2026},
      booktitle = {The Web Conference}
}

@inproceedings{0001K0B26,
  author       = {A. Khan and
                  X. Ke and
                  Y. Wu and
                  F. Bonchi},
  title        = {{GNN} Explainers 2.0: User-centric and Data-driven Insights},
  booktitle    = {WSDM},
  pages        = {1359--1362},
  year         = {2026}
}

@inproceedings{yuan2021explainability,
  title={On explainability of graph neural networks via subgraph explorations},
  author={Yuan, H. and Yu, H. and Wang, J. and Li, K. and Ji, S.},
  booktitle={ICML},
  year={2021}
}

@inproceedings{ying2019gnnexplainer,
  title={Gnnexplainer: Generating explanations for graph neural networks},
  author={Ying, Z. and Bourgeois, D. and You, J. and Zitnik, M. and Leskovec, J.},
  booktitle={NeurIPS},
  year={2019}
}

@inproceedings{YuanTHJ20,
  author       = {H. Yuan and
                  J. Tang and
                  X. Hu and
                  S. Ji},
  title        = {XGNN: Towards model-level explanations of graph neural networks},
  booktitle    = {KDD},
  year         = {2020}
}

@article{10.1145/3749116.3749122,
author = {Khan, A. and Ke, X. and Wu, Y.},
title = {Graph Data Management and Graph Machine Learning: Synergies and Opportunities},
year = {2025},
volume = {54},
number = {2},
journal = {SIGMOD Rec.},
pages = {28–42}
}

@article{10.1145/3626246.3654735,
  author = {Chen, T. and Qiu, D. and Wu, Y. and Khan, A. and Ke, X. and Gao, Y.},
  title = {View-based Explanations for Graph Neural Networks},
  year = {2024},
  journal      = {Proc. {ACM} Manag. Data},
  volume       = {2},
  number       = {1},
  pages        = {40:1--40:27}
}

\end{document}